\documentclass[runningheads]{llncs}
\usepackage[T1]{fontenc}
\usepackage{graphicx}
\usepackage{amsmath}
\usepackage{enumitem}
\usepackage{multirow}
\usepackage{url}
\usepackage{booktabs}

\begin{document}
\title{Is the Statistical Advantage Worth the Cost? An Empirical Comparison of KANs and MLPs for Structured Data Classification}
\titlerunning{Empirical Comparison of KANs and MLPs for Structured Data}
% If the paper title is too long for the running head, you can set
% an abbreviated paper title here
%
\author{
    Matthew Steven P. Toledo
    \inst{1,2}
    %\orcidID{0009-0006-9974-7711} 
    \and
    Justine Raphael H. Jacinto
    \inst{1,2}
    %\orcidID{0009-0002-2644-8535} 
    \and
    Vivekjeet Singh Chambal
    \inst{1,2}
    \and
    Rodolfo C. Camaclang III
    \inst{1,2}
    \and
    Jamlech Iram N. Gojo Cruz
    \inst{1,2}
    \and
    Reginald Neil C. Recario
    \inst{1,2}
}
\authorrunning{M. S. P. Toledo et al.}
% First names are abbreviated in the running head.
% If there are more than two authors, 'et al.' is used.
%
% \institute{Institute of Computer Science, University of the Philippines Los Baños, Philippines
% \email{sampleemail1@up.edu.ph}\\
% \url{http://www.springer.com/gp/computer-science/lncs} \and
% ABC Institute, Rupert-Karls-University Heidelberg, Heidelberg, Germany\\
% \email{\{abc,lncs\}@uni-heidelberg.de}}

\institute{
Institute of Computer Science, University of the Philippines Los Baños, Philippines\\
\and Machine Learning and Artificial Intelligence Applications Lab, University of the Philippines Los Baños, Philippines\\
\email{\{mptoledo2, jhjacinto2, vcchambal, rccamaclang1, jngojocruz, rcrecario\}@up.edu.ph} 
}

% wala naman tayong website lol

%
\maketitle              % typeset the header of the contribution
\begin{abstract}
This study presents an empirical benchmarking comparison between Kolmogorov-Arnold Networks (KANs) and Multi-Layer Perceptrons (MLPs) on structured tabular classification tasks. Motivated by the growing interest in KANs as an alternative function-approximating architecture, we evaluate their out-of-the-box performance on twelve publicly available datasets spanning binary, multiclass, multilabel, and ordinal problems. Both models were trained under standardized preprocessing, architecture, and fixed hyperparameter settings, with performance assessed using test accuracy and F1-Score, paired hypothesis testing, and effect size analysis. Results show that KANs statistically outperform MLPs in binary and multiclass domains and achieve a significant aggregate advantage across all datasets. However, the observed medium effect size ($d = -0.46$) raises an important cost-benefit consideration: while KANs offer superior generalization through adaptive spline-based mappings, this advantage comes with substantially higher parameter and computational complexity relative to the MLP baseline. These findings suggest KANs are the preferred choice for high-precision applications, while MLPs remain a robust and efficient option for resource-constrained environments. Future work should extend this analysis to additional data modalities to further refine these architectural selection criteria.

\end{abstract}

\section{Introduction}
Structured tabular data remains one of the most widely used input formats in production machine learning systems, spanning fields from healthcare to finance. In these settings, accuracy, reliability, and training efficiency are critical, and the choice of model architecture often has a direct influence on practical decisions \cite{1_treebased_grinsztajn}. Despite continued advances in deep learning, no single architecture has established itself as clear and consistent best choice for tabular classification \cite{2_survey_borisov,3_revisiting_gorishniy}, and the trade-off between model expressivity and computational cost remains an ongoing concern for practitioners \cite{5_tabular_shwartz}.

Recent advances have introduced Kolmogorov-Arnold Networks (KAN) as a theoretically motivated alternative to the multilayer perceptron (MLP), offering a fundamentally different approach to function approximation \cite{4_kan_liu}. While early results have been promising, whether this architectural difference translates to a meaningful and computationally justified performance advantage on real-world tabular tasks remains an open empirical question. Most existing comparisons are limited in scope and vary widely in experimental conditions, making general conclusions difficult \cite{4_kan_liu,20_benchmark_poeta}

This study addresses that gap by empirically comparing KANs and MLPs across a diverse set of structured tabular classification tasks under standardized experimental settings. In doing so, it not only evaluates raw performance differences but also examines their statistical significance and practical effect size, and weighs these gains against the additional computational cost of KANs. Specifically, the study aims to:
\begin{enumerate}
\item Implement and train KAN and MLP models with comparable architectures and fixed, untuned hyperparameters across different classification task types (binary, multiclass, multilabel, and ordinal);
\item Evaluate and compare their performance using test accuracy and F1-Score as the primary metrics;
\item Determine the statistical significance of the observed performance differences between KAN and MLP using paired hypothesis tests; and
\item Analyze the aggregate and per-task trends to weigh the predictive gains against architectural complexity.
\end{enumerate}

\section{Background and related works}
Structured tabular data remain a dominant input format in machine learning and are widely used in domains such as healthcare, education, finance and remote sensing \cite{6_cancelout_borisov}. These datasets are challenging to model because they often contain heterogeneous feature types, missing values, nonlinear relationships, and limited sample sizes. As a result, traditional decision tree ensemble methods such as Random Forests and Gradient Boosted Trees continue to be the preferred choice for many tabular learning problems since they handle feature heterogeneity and implicit feature interactions effectively with minimal preprocessing requirements \cite{7_simple_kadra}. However, while these methods remain competitive, the growing interest in neural architectures that offer expressive modeling, scalability, and integration into deep learning pipelines has renewed attention on feed-forward networks as a practical alternative for tabular tasks.

Among feed-forward architectures, the multilayer perceptron remains the most common and classical architecture \cite{8_mlp_przybyla}. An MLP learns by applying successive linear transformations and fixed nonlinear activation functions across hidden layers and optimizing parameters through gradient based methods \cite{9_backprop_rumelhart,10_dl_goodfellow}. Its theoretical universality and strong empirical performance have established it as a standard baseline in many studies and benchmarks \cite{11_universal_hornik,12_embeddings_gorishniy,13_tabnet_arik,14_saint_somepalli}. However, the use of fixed activation functions and scalar weights can limit the ability of MLPs to capture localized nonlinear effects without increasing network depth or width, which can lead to higher computational cost and tuning complexity \cite{15_nn_haykin}. These limitations have driven research toward adaptive activation mechanisms and function-based representations that learn nonlinear transformations directly from data to improve flexibility and interpretability \cite{16_spline_bohra,17_lipschitz_ducotterd}.

One such direction is Kolmogorov Arnold Networks, which are inspired by the Kolmogorov-Arnold theorem on multivariate function representation \cite{18_kolmogorov1956,19_arnold1957}. The theorem demonstrates that any continuous multivariate function can be written as the finite sum of simpler univariate continuous functions, implying that complex high-dimensional relationships can theoretically be constructed from simpler one-dimensional transformations. Each connection in the network corresponds to a small parametric function that often uses spline interpolation to learn how inputs should transform along individual dimensions. KANs replace fixed activations with learnable univariate functions on network connections, which shifts expressivity from node activations to edge-based mappings and enables localized nonlinear modeling. This fundamental architectural departure from the MLP makes a direct empirical comparison between the two a natural and important research question.

Empirical studies have shown that KANs can match or exceed MLP performance on tabular and scientific datasets while offering improved interpretability and faster convergence in some cases \cite{4_kan_liu,20_benchmark_poeta}. Existing benchmarks vary widely in datasets, hyperparameter choices, and task coverage, which makes general conclusions difficult \cite{4_kan_liu,20_benchmark_poeta,21_genomics_zhang}. Most prior work focuses on binary and multiclass classification, while multilabel and ordinal settings remain largely unexplored \cite{21_genomics_zhang}. This gap limits understanding of how KANs generalize across classification paradigms and motivates systematic evaluation under standardized conditions to determine whether KANs offer a consistent and cost-justified advantage over MLPs across a broad range of tabular classification tasks, particularly in domains such as healthcare decision support where interpretability and reliability are critical \cite{1_treebased_grinsztajn,22_explainability_holzinger}.

\section{Methods}

\subsection{Datasets and task categories}
This study benchmarks the performance of KANs against conventional MLPs across four classification paradigms: binary, multiclass, multilabel, and ordinal. To ensure a rigorous and reproducible evaluation, twelve datasets were selected from publicly available repositories, including the University of California, Irvine (UCI) Machine Learning Repository, Kaggle, and MULAN library for multi-label learning. These datasets were chosen to represent a diverse cross-section of structured tabular data, encompassing varying sample sizes, feature modalities, and problem domains. These domains range from healthcare and education to biology and physical sciences. Accordingly, the selected datasets are categorized into four primary task types, as shown in Table~\ref{tab:datasets}.

\begin{table}[ht]
\centering
\caption{Summary of Datasets Used for Evaluation}
\label{tab:datasets}
\begin{tabular}{llrp{5.5cm}c}
\hline\noalign{\smallskip}
\textbf{Task} & \textbf{Dataset} & \textbf{$N$} & \textbf{Description} & \textbf{Ref.} \\
\noalign{\smallskip}\hline\noalign{\smallskip}
\multirow{3}{*}{Binary} & Employability (Kaggle) & 2,982 & Predict student interview outcomes & \cite{kaggle_employability} \\
 & AIDS Clinical (UCI) & 2,139 & Patient survival prediction & \cite{uci_aids} \\
 & Sec. Mushroom (UCI) & 61,069 & Edible vs. poisonous classification & \cite{uci_mushroom} \\
\hline
\multirow{3}{*}{Multiclass} & Student Dropout (UCI) & 4,424 & Dropout, enrolled, or graduate & \cite{uci_dropout,uci_dropout_extra} \\
 & Yeast (UCI) & 1,484 & Protein localization sites & \cite{uci_yeast} \\
 & Statlog Satellite (UCI) & 6,435 & Land cover classification & \cite{uci_statlog} \\
\hline
\multirow{3}{*}{Multilabel} & Emotions (MULAN) & 593 & Music mood classification & \cite{mulan_emotions} \\
 & Birds (MULAN) & 645 & Audio-based species detection & \cite{mulan_birds} \\
 & Enron (MULAN) & 1,702 & Email text categorization & \cite{mulan_enron} \\
\hline
\multirow{3}{*}{Ordinal} & Balance Scale (UCI) & 625 & Psychological balance modeling & \cite{uci_balance_scale} \\
 & Car Evaluation (UCI) & 1,728 & Car acceptability rating & \cite{uci_car_evaluation} \\
 & Abalone (UCI) & 4,177 & Age prediction via physical metrics & \cite{uci_abalone} \\
\hline
\end{tabular}
\end{table}

\subsection{Data preprocessing}
A uniform preprocessing pipeline was applied across all datasets. Missing values were handled through listwise deletion or imputation using the mean for numerical and mode for categorical features. Nominal attributes underwent one-hot encoding, ordinal features retained their inherent order, and all numerical features were standardized to zero mean and unit variance. Target variables were processed according to task type, with multilabel targets retained as binary indicator vectors. Datasets were partitioned using stratified sampling into 70\% training, 15\% validation, and 15\% testing sets for standard tasks, and an 80-20 train-test split for multilabel tasks due to the computational complexity of stratifying multi-label data.

\subsection{Model configuration and evaluation}
To ensure comparability and fairness, each dataset undergoes identical preprocessing, model configuration, and training conditions, with experiments conducted in uniform hardware environments except where computational constraints required deviation. This standardized pipeline was intentionally kept fixed to evaluate each architecture as-is, without tuning, ensuring that any observed differences reflect architectural properties rather than optimization effort. To ensure a meaningful comparison without the confounding variable of tuning, both models utilized "vanilla" configurations established in foundational literature, summarized in Table~\ref{tab:model_config}. These configurations isolate the difference between fixed node-based activations and adaptive spline mappings while maintaining equivalent parameter complexity. 

\begin{table}[h]
\centering
\caption{Summary of model architecture and training configuration.}
\label{tab:model_config}
\renewcommand{\arraystretch}{1.2}
\begin{tabular}{lll}
\hline
\textbf{Configuration} & \textbf{MLP} & \textbf{KAN} \\
\hline
Reference & Gorishniy et al.~\cite{12_embeddings_gorishniy} & Liu et al.~\cite{4_kan_liu} \\
Hidden layers (standard)$^*$ & [16, 16] & [16, 16] \\
Hidden layers (multilabel)$^*$ & [128, 64] & [64, 32] \\
Activation & ReLU & Spline (learnable) \\
Dropout & $p = 0.2$ & --- \\
Grid size / order (standard) & --- & Grid 5, $k = 3$ \\
Grid size / order (multilabel) & --- & Grid 3, $k = 2$ \\
Optimizer & Adam & Adam \\
Learning rate & $1 \times 10^{-3}$ & $1 \times 10^{-3}$ \\
Loss (binary / multilabel) & BCE with Logits & BCE with Logits \\
Loss (multiclass) & Cross-Entropy & Cross-Entropy \\
Loss (ordinal) & Ordinal Cross-Entropy & Ordinal Cross-Entropy \\
Epochs (standard) & 100 & 100 \\
Epochs (multilabel) & 20 & 20 \\
Batch size (large datasets) & 256 & 256 \\
\hline
\multicolumn{3}{l}{$^*$\footnotesize{Input and output dimensions are prepended and appended respectively.}}\\
\end{tabular}
\end{table}

Performance was evaluated using Accuracy, to measure overall predictive correctness, and F1-Score, to provide a balanced assessment of precision and recall, with calculation strategies tailored to each classification paradigm. For binary, multiclass, and ordinal tasks, weighted F1-Score was used to account for class imbalance, while multilabel tasks used subset accuracy and sample-averaged F1-Score. All models underwent five independent execution runs, and average metrics are reported. To confirm that performance differences represent a systematic architectural advantage rather than stochastic noise, paired hypothesis testing was conducted at $\alpha = 0.05$, with test selection governed by the Anderson-Darling normality test. Effect sizes were reported using Cohen's \(d\) for t-tests and rank-biserial correlation ($r_{rb}$) for the Wilcoxon signed-rank test.

\section{Results and discussion}

\subsection{Computational Efficiency and Resource Requirements}
Shown in Table~\ref{tab:kan_mlp_params_time} are the summary for the parameter count, training time, and inference time of both KAN and MLP for each dataset. The data consistently put KAN as more computationally "expensive" than MLP.

\begin{table}[h]
\caption{Summary of parameter count and average computational time for MLP and KAN models across 12 benchmark datasets grouped by classification type.}
\label{tab:kan_mlp_params_time}
\centering
\resizebox{\textwidth}{!}{% <-- Added resizebox here
\begin{tabular}{@{}llrrrrrr@{}}
\toprule
\textbf{Category} & \textbf{Dataset} &
\multicolumn{2}{c}{\textbf{Parameter Count}} &
\multicolumn{2}{c}{\textbf{Train Time (s)}} &
\multicolumn{2}{c}{\textbf{Inference (ms)}} \\
\cmidrule(lr){3-4} \cmidrule(lr){5-6} \cmidrule(lr){7-8}
 &  & \textbf{MLP} & \textbf{KAN} & \textbf{MLP} & \textbf{KAN} & \textbf{MLP} & \textbf{KAN}  \\
\midrule
Binary
& Employability        & \textbf{433}    & 7012   & \textbf{0.93} & 48.28  & \textbf{0.98}  & 59.53   \\
& AIDS Clinical Trials & \textbf{673}   & 11032  & \textbf{0.58} & 61.41  & \textbf{0.78} & 94.84  \\
& Mushrooms            & \textbf{2209} & 36760  & \textbf{1.96} & 628.57 & \textbf{2.77} & 1117.31 \\
\midrule
Multiclass
& Dropout              & \textbf{915}   & 15036  & \textbf{0.64} & 83.35  & \textbf{0.60} & 151.28 \\
& Yeast                & \textbf{586}   & 9352   & \textbf{0.59} & 55.50  & \textbf{1.19}  & 74.53   \\
& Statlog (Satellite)  & \textbf{966}   & 15816  & \textbf{1.41} & 124.17 & \textbf{1.60} & 220.97\\
\midrule
Multilabel
& Emotions             & \textbf{17990}   & 75328   & \textbf{1.33} & 1128.14  & \textbf{0.01}   & 5.50  \\
& Birds                & \textbf{42899}   & 212256  & \textbf{1.00} & 1587.21 & \textbf{0.67}   & 22.14   \\
& Enron                & \textbf{139957}   & 745888  & \textbf{3.09} & 7101.09 & \textbf{0.15}   & 32.96  \\
\midrule
Ordinal
& Balance Scale        
& \textbf{403}    & 6460   & \textbf{0.55} & 30.35  & \textbf{0.75}   & 43.63  \\
& Car Evaluation       & \textbf{452}   & 7256   & \textbf{0.44} & 42.25  & \textbf{1.42}   & 80.80   \\
& Abalone             
& \textbf{501}    & 8052  & \textbf{1.04} & 57.16  & \textbf{1.73}   & 88.02  \\
\bottomrule
\end{tabular}%
} % <-- Closing brace for resizebox
\end{table}

In terms of architectural complexity, MLP requires only a single scalar weight per connection, resulting in a parameter complexity of \(O(N^2L)\), where \(N\) is the network width and \(L\) is the depth. KAN replaces each weight with a learnable spline function, resulting in a parameter complexity of \(O(N^2GL)\) and a computational complexity of \(O(2kBN^2GL)\) for a training batch of size \(B\), where \(G\) is the grid size and \(k\) is the spline order \cite{4_kan_liu}. The additional factor of \(GL\) over the MLP in parameter complexity directly explains the substantially higher parameter counts observed in Table~\ref{tab:kan_mlp_params_time}, where KAN consistently requires roughly 16 times more parameters than MLP across standard tasks.

Furthermore, the computational complexity factor of \(2k\), arising from the recursive evaluation of order-\(k\) splines, accounts for the disproportionately higher training and inference times: KAN training times exceed 30 seconds across all datasets while MLP training rarely exceeds five seconds, and KAN inference latency frequently exceeds 100 ms while MLP remains 3 ms. Together, these therotical complexities provide a principled explanation for the computational overhead observed empirically, and directly motivate the cost-benefit analysis presented in Section 4.4.

\begin{figure}[th!]
    \centering
    \includegraphics[width=\linewidth]{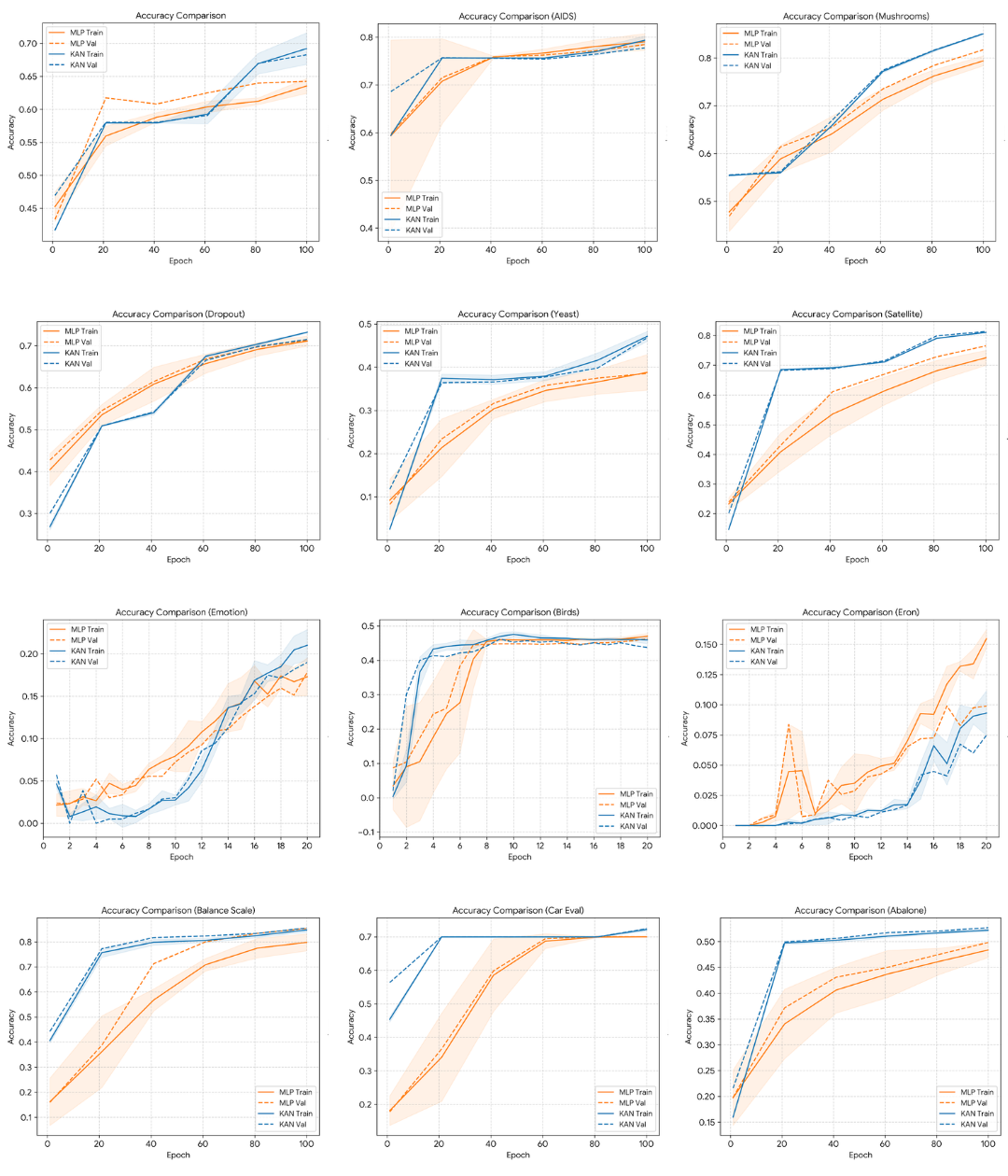}
    \caption{From Top to Bottom: Training and validation accuracy of KAN and MLP across dataset categories from Binary (top row), Multiclass (middle row), Multilabel (third row), and Ordinal (bottom row) class tasks.}
    \label{fig:kan_mlp_summary}
\end{figure}

\subsection{Performance Evaluation: Accuracy and F1-Score}

Shown in Figure~\ref{fig:kan_mlp_summary} are the training and validation accuracy of both KAN and MLP per dataset. For the majority of the datasets, both have a steady improvement for both training and validation accuracy and reach for similar upper limits, except for \textit{Emotions}, \textit{Birds}, and \textit{Enron} datasets. However, KAN reaches the limits faster and stabilizes earlier than MLP for most of the dataset, except for \textit{Emotions}, \textit{Birds}, and \textit{Enron} datasets. 

Table~\ref{tab:kan_mlp_testaccuracy_f1score} presents the evaluation results of the KAN and MLP models across all datasets.  In terms of test accuracy, among the three datasets in both binary and multiclass classification, KAN has a higher test accuracy than MLP. For the multilabel classification, KAN has a higher test accuracy than MLP for one dataset only, while MLP is better on the birds and enron datasets. Meanwhile, KAN has a higher test accuracy for two out of three datasets in the ordinal classification. In terms of F1-Score, KAN is consistently higher than MLP on all datasets under binary and multiclass classification. For multilabel classification, KAN has a higher F1-Score than MLP only on two datasets, except for \textit{Enron} dataset. Lastly, for ordinal classification, KAN has a higher F1-Score on two out of three datasets.

\begin{table}[h]
\caption{Summary of test accuracy and F1-score for MLP and KAN models across 12 benchmark datasets grouped by classification type.}
\label{tab:kan_mlp_testaccuracy_f1score}
\centering
\setlength{\tabcolsep}{7pt} % tighter column spacing
\begin{tabular}{@{}llrrrr@{}}
\toprule
\textbf{Category} & \textbf{Dataset} &
\multicolumn{2}{c}{\textbf{Accuracy}} &
\multicolumn{2}{c}{\textbf{F1-Score}} \\
\cmidrule(lr){3-4} \cmidrule(lr){5-6}
 &  & \textbf{MLP} & \textbf{KAN} & \textbf{MLP} & \textbf{KAN}  \\
\midrule
Binary
& Employability        & 0.6214 & \textbf{0.6694} & 0.3482 & \textbf{0.5647} \\
& AIDS Clinical Trials & 0.7888 & \textbf{0.7933} & 0.2887 & \textbf{0.3310} \\
& Mushrooms            & 0.8177 & \textbf{0.8513} & 0.8338 & \textbf{0.8653} \\
\midrule
Multiclass
& Dropout              & 0.7130 & \textbf{0.7154} & 0.6410 & \textbf{0.6496} \\
& Yeast                & 0.3946 & \textbf{0.4760} & 0.3151 & \textbf{0.4210} \\
& Statlog (Satellite)  & 0.7700 & \textbf{0.8162} & 0.7140 & \textbf{0.7743} \\
\midrule
Multilabel
& Emotions             & 0.1782 & \textbf{0.1899} & 0.3778 & \textbf{0.4373} \\
& Birds                & \textbf{0.4636} & 0.4372 & 0.0267 & \textbf{0.0280} \\
& Enron                & \textbf{0.0909} & 0.0639 & \textbf{0.4750} & 0.4309 \\
\midrule
Ordinal
& Balance Scale        & \textbf{0.8586} & 0.8484 & \textbf{0.8246} & 0.8148 \\
& Car Evaluation       & 0.7014 & \textbf{0.7185} & 0.5794 & \textbf{0.6208} \\
& Abalone              & 0.4917 & \textbf{0.5217} & 0.4116 & \textbf{0.4700} \\
\bottomrule
\end{tabular}
\end{table}

Overall, in terms of test accuracy, KAN performed better than MLP for 9 out of 12 datasets while MLP performed better than KAN in 3 datasets. In terms of the F1-Score, KAN performed better than MLP for 10 out of 12 datasets, while MLP performed better in the remaining 2 datasets. This means that across the range of the datasets tested, KAN performed better than MLP in the majority of datasets.

\subsection{Statistical Significance and Effect Size}

While comparing average metrics reveals performance trends, it does not account for the variance inherent in model training. To confirm that the observed differences represent a systematic architectural advantage rather than stochastic noise, we subjected the results to formal hypothesis testing on test accuracy. Table~\ref{tab:stat_results} shows the summary of the statistical comparison.

\begin{table}[h]
    \centering
    \caption{Summary of Statistical Comparison of Average Test Accuracy between MLP and KAN Architectures}
    \label{tab:stat_results}
    \begin{tabular}{lcccccc}
    \toprule
    \textbf{Category} & \textbf{Stat. Test} & \textbf{Mean Diff.} & \textbf{$p$-value} & \textbf{Effect Size} & \textbf{Result} \\
     & & \small{(MLP - KAN)} & & & \\ \hline
    \textbf{Binary} & Paired $t$-Test & -0.029 & \textbf{0.001} & $d = -1.09$ & \textbf{KAN Superior} \\
    \textbf{Multiclass} & Wilcoxon & -0.039\textsuperscript{$\dagger$} & \textbf{0.001} & $r_{rb} = -1.00$ & \textbf{KAN Superior} \\
    \textbf{Multilabel} & Paired $t$-Test & +0.014 & 0.066 & $d = +0.51$ & No Sig. Diff. \\
    \textbf{Ordinal} & Paired $t$-Test & -0.012 & 0.111 & $d = -0.44$ & No Sig. Diff. \\ \hline
    \textbf{Overall} & \textbf{Paired $t$-Test} & \textbf{-0.018} & \textbf{0.001} & \textbf{$d = -0.46$} & \textbf{KAN Superior} \\\bottomrule
    \end{tabular}
    \vspace{0.2cm}
    \footnotesize{\\ \textsuperscript{$\dagger$}For Wilcoxon test, value represents the Estimated Median Difference.}
\end{table}

For the analysis per classification type, the statistical tests revealed task-dependent outcomes. Specifically, for Binary and Multiclass datasets, the tests showed $p$-values less than $\alpha = 0.05$, indicating a significant difference where KAN outperformed MLP. However, for Multilabel and Ordinal tasks, there is no statistically significant difference between the performance of KAN and MLP for these specific metrics. While KAN demonstrates clear superiority in binary and multiclass domains, the two models perform similarly in multilabel and ordinal contexts. 

The analysis of effect sizes reveals that KAN demonstrates a large, decisive performance advantage in Binary ($d=-1.09$) and Multiclass ($r_{rb}=-1.00$) tasks. In contrast, the results for Multilabel and Ordinal tasks were less conclusive, showing only medium-to-small effects ($d=0.51$ favoring MLP and $d=-0.44$ favoring KAN, respectively) that did not reach statistical significance.
    
In contrast, for the aggregate analysis across all datasets, the test showed that there is a significant difference ($p=0.001$) between the performance of KAN and MLP in favor of KAN. With a medium effect size ($d = -0.46$), this confirms that KAN generally outperforms MLP across the broad spectrum of classification tasks tested.

\subsection{Cost-Benefit Analysis: Accuracy vs. Efficiency}
    
Despite the statistical significance of KAN's performance advantage, the practical implication of a "medium" effect size ($d = -0.46$) warrants a critical cost-benefit analysis. KAN may have demonstrated a generalized superiority, but the margin of improvement is not transformative enough to think MLPs obsolete, particularly when considering computational constraints. The $B$-spline computations inherent to KANs typically incur a higher computational overhead compared to the highly optimized matrix multiplications of standard MLPs.

Consequently, the choice between these architectures should be dependent on context. For high-stakes applications where predictive precision is extremely crucial, such as in medical diagnosis or financial forecasting, the statistical advantage of KAN justifies the additional computational cost. However, for environments with resource constraints, such as real-time edge computing or low-latency systems, the "medium" performance gain may not outweigh the efficiency losses. In such scenarios, the MLP remains a robust and computationally economical baseline. Thus, while KAN represents a theoretically superior architecture for these datasets, its adoption should be driven by the specific tolerance for computational overhead versus the need for marginal accuracy gains.

\section{Conclusion}

This study presented an empirical comparison between Kolmogorov-Arnold Networks (KANs) and Multi-Layer Perceptrons (MLPs) on structured data classification tasks. Using twelve datasets spanning binary, multiclass, multilabel, and ordinal problems, both models were evaluated in terms of test accuracy and F1-Score under standardized training conditions. 

Aggregate results confirm that KAN statistically outperforms MLP, though the performance gap varies by classification type. The magnitude of this improvement,quantified as a medium effect size, poses a pivotal cost-benefit question. The decision to adopt KAN should therefore be driven by the specific operational requirement: prioritize KAN for maximum accuracy, but retain MLP when computational efficiency is the primary constraint.

Beyond empirical results, the findings highlight KAN’s flexibility in adapting to complex tabular patterns through its spline-based functional mappings, which allow it to approximate nonlinear relationships more efficiently. These characteristics point to KAN as a promising direction for structured data modeling, offering a balance between performance and computational efficiency.

% As this study focused exclusively on structured tabular data, future work should explore broader domains such as text, images, and time series to evaluate the generality of KAN’s advantages. Further research may also include deeper theoretical analyses, and hyperparameter optimization to improve the robustness and interpretability of the results.

Several limitations of this study should be acknowledged. The experiments were conducted under fixed hyperparameters and architectural configurations to fairly compare both models in their default state, which improves comparability but may not fully capture each model's optimal performance under extensive tuning. Additionally, multilabel datasets were trained for fewer epochs ($i = 20$) than other task types ($i = 100$) due to computational constraints, which may have limited convergence for more complex models such as KAN. Future work should extend this analysis to additional data modalities including text, images, and time series, and may include deeper theoretical analyses and hyperparameter optimization to improve the robustness and interpretability of the results.

% \section{other stuff (to delete/edit)}

% % ======================
% % ======================
% % ======================
% % ======================

% % citations by square brackets nalang tayo
% %below is from the template
% For citations of references, we prefer the use of square brackets
% and consecutive numbers. Citations using labels or the author/year
% convention are also acceptable. 

% % The following bibliography provides
% % a sample reference list with entries for journal
% % articles~\cite{ref_article1}, an LNCS chapter~\cite{ref_lncs1}, a
% % book~\cite{ref_book1}, proceedings without editors~\cite{ref_proc1},
% % and a homepage~\cite{ref_url1}. Multiple citations are grouped
% % \cite{ref_article1,ref_lncs1,ref_book1},
% % \cite{ref_article1,ref_book1,ref_proc1,ref_url1}.

% \begin{credits}
% \subsubsection{\ackname}
% The authors would like to thank Sir Rodolfo C. Camaclang for suggesting the initial research direction on Kolmogorov–Arnold Networks, and Sir Jamlech Iram Gojo Cruz for his guidance in shaping the empirical comparison design. The authors also thank Sir Reginald Neil C. Recario for his valuable feedback and support during the research process.

% \subsubsection{\discintname}
% The authors have no competing interests to declare that are relevant to the content of this article.
% \end{credits}
%
% ---- Bibliography ----
%
% BibTeX users should specify bibliography style 'splncs04'.
% References will then be sorted and formatted in the correct style.
%
\bibliographystyle{splncs04}
\bibliography{references}
\end{document}